%% file: root.tex
\documentclass[letterpaper, 10 pt, conference]{ieeeconf}

\IEEEoverridecommandlockouts                              

\usepackage{graphicx}
\usepackage{color}
\usepackage{xcolor}
\usepackage{adjustbox}
\usepackage{amssymb}
\usepackage{pgfplots}
\usepackage{subcaption}

\usepackage{microtype} 

\pgfplotsset{
every axis/.append style={
  axis line style={->}, 
  legend style={font=\scriptsize},
  label style={font=\scriptsize},
  tick label style={font=\scriptsize},
  xlabel={$x$},          
  ylabel={$y$},          
  }
}

\usepackage{hyperref}
\usepackage{booktabs,multirow,tabularx}

\usetikzlibrary{positioning}
\usetikzlibrary{decorations.pathreplacing}
\usetikzlibrary{calc}
\usetikzlibrary{positioning,shapes,shadows,arrows}
\usepackage{booktabs, dcolumn}
\usepackage{siunitx}

\usepackage[english]{babel}
\usepackage{lipsum}

\usepackage{pgfplots}
\pgfplotsset{compat=newest}

\usepackage{amsmath}

\usepackage[vlined,ruled]{algorithm2e}
\usepackage{url}
\usepackage{mathabx}
\usepackage{mathrsfs}
\usepackage{ntheorem}
\theoremheaderfont{\bfseries \itshape}
\theoremseparator{:}

\makeatletter
\let\NAT@parse\undefined
\makeatother

\usepackage[numbers,sort,compress]{natbib}
\hypersetup{draft}

\usepackage[font=small]{caption}

\newcommand{\secref}[1]{Sec.~\ref{#1}}
\newcommand{\figref}[1]{Fig.~\ref{#1}}
\newcommand{\algref}[1]{Algorithm~\ref{#1}}
\renewcommand{\eqref}[1]{Eq.~(\ref{#1})}

\newcommand{\tableref}[1]{Table~\ref{#1}}

\title{\bf Multi-Object Rearrangement with Monte Carlo Tree Search:\\ A Case Study on Planar Nonprehensile Sorting}

\author{
Haoran Song$^{1*}$, Joshua A. Haustein$^{2*}$, Weihao Yuan$^{1}$, \\
Kaiyu Hang$^{3}$, Michael Yu Wang$^{1}$, Danica Kragic$^{2}$, Johannes A. Stork$^{4}$
\thanks{$^{1}$H. Song, W. Yuan and M. Y. Wang are with the Robotics Institute, Hong Kong University of Science and Technology, Hong Kong, China.}
\thanks{$^{2}$J. A. Haustein and D. Kragic are with the Centre for Autonomous Systems, EECS, KTH Royal Institute of Technology, Stockholm, Sweden.}
\thanks{$^{3}$K. Hang is with the Department of Mechanical Engineering and Material Science, Yale University, New Haven, Connecticut, USA}
\thanks{$^{4}$J. A. Stork is with the Centre for Applied Autonomous Sensor Systems, Orebro University, Orebro, Sweden.}
\thanks{$^{*}$These authors contributed equally to this work}
}

\begin{document}

\maketitle
\thispagestyle{empty}
\pagestyle{empty}

\input{abstract}
\input{introduction}
\input{problem-def}
\input{relatedwork}
\input{method}
\input{method-mcts}
\input{method-policy}
\input{experiments}
\input{conclusion}





\footnotesize
\bibliography{references}

\end{document}

%% file: abstract.tex

\begin{abstract}
In this work, we address a planar non-prehensile sorting task.
Here, a robot needs to push many densely packed objects belonging to different classes into a configuration where these classes are clearly separated from each other.
To achieve this, we propose to employ Monte Carlo tree search equipped with a task-specific heuristic function.
We evaluate the algorithm on various simulated and real-world sorting tasks.
We observe that the algorithm is capable to reliably sort large numbers of convex and non-convex objects, as well as convex objects in the presence of immovable obstacles.
\end{abstract}

%% file: introduction.tex

\section{INTRODUCTION}
Rearranging objects, e.g., to clear a path or clean a table, is an essential skill for an autonomous robot. The robot needs to plan in which order to move the objects and whereto.
This rearrangement planning problem is known to be NP- or even PSPACE-hard depending on the goal definition~\cite{Wilfongpspace}.
Accordingly, various specialized algorithms have been proposed that address specific practical rearrangement problems efficiently.
For instance, several prior works specifically address navigating a mobile robot among movable obstacles (NAMO)~\cite{Stilman2005,Stilman2008,VandenBerg2009,Nieuwenhuisen2008}.
Similarly, clearing clutter for grasping has been addressed by pushing obstacles aside locally~\cite{Kitaev2015,Agboh2018,Muhayyuddin2018}, or recursively removing obstructions through pick-and-place~\cite{Stilman2007}.
Even for large-scale rearrangements, where many objects need to be arranged to target locations, efficient approximative algorithms have been proposed.
While early works~\cite{Ben-Shahar1998b,Stilman2007} were limited to \textit{monotone} problems, where each object needs to be moved at most once, recent works have overcome this limitation~\cite{Krontiris2015,Krontiris2016,Garrett2018,Han2017,Huang2019}.
Large-scale rearrangements, however, have predominantly been addressed using pick-and-place or single-object pushing.

\begin{figure}[t]
    \centering
    \includegraphics[width = 0.98 \columnwidth]{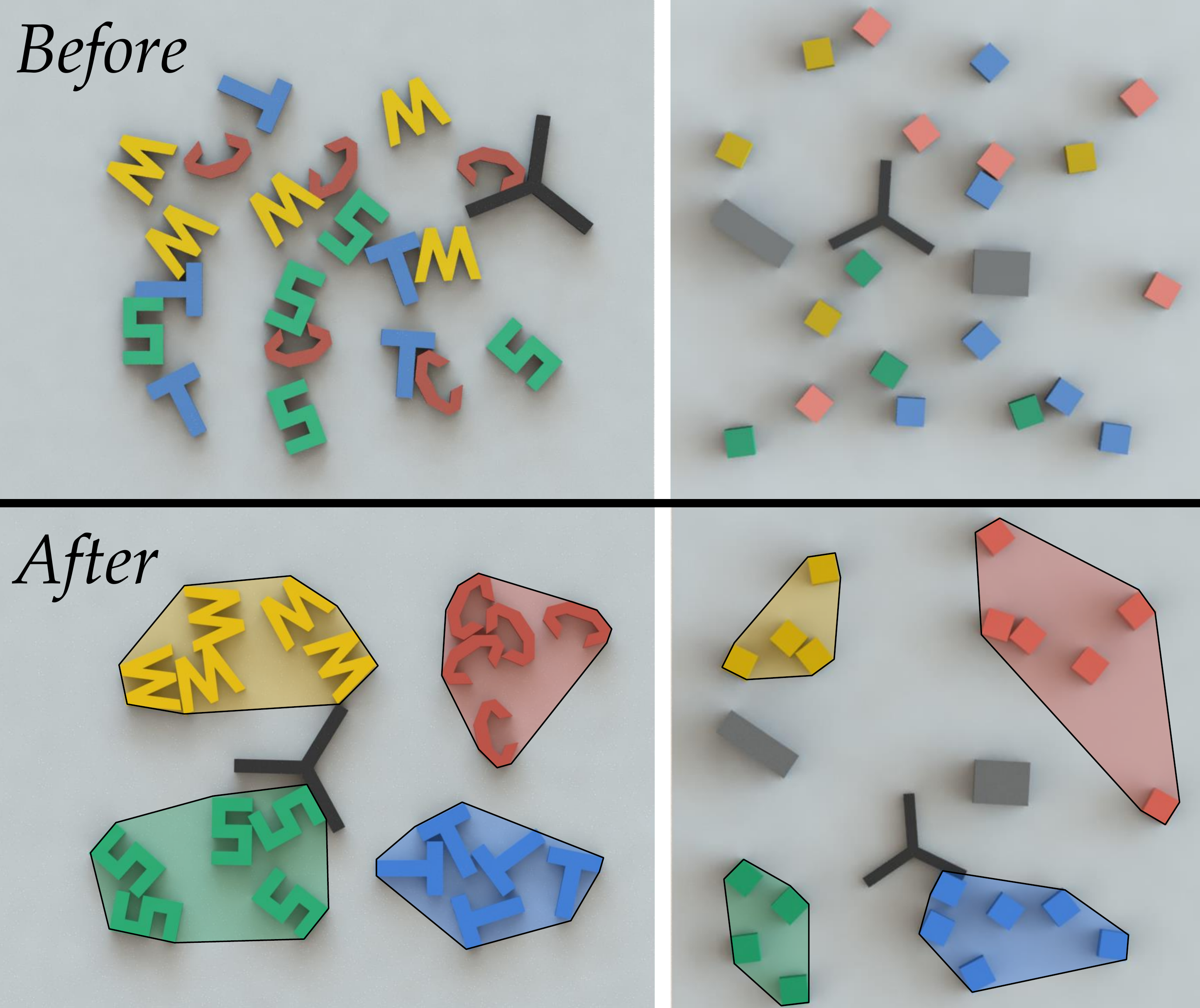}
    \caption{The planar push sorting task: A planar robot (black) is tasked to separate
        objects belonging to different classes into homogenous distinct clusters,
        optionally in the presence of obstacles (grey).}
\label{fig:problem}
\end{figure}

We are interested in large-scale \emph{non-prehensile} rearrangement problems, where a robot pushes multiple objects simultaneously to reach a goal that is characterized by the final poses of many objects.
Specifically, we consider the planar non-prehensile sorting task illustrated in \figref{fig:problem}.
Here, a planar pusher has to sort objects belonging to different classes into homogenous distinct clusters. The problem is challenging, as it is non-monotone, requires multi-object pushing to be solved efficiently, and involves robot motion planning to circumnavigate obstacles.

We propose to address this problem with Monte Carlo tree search (MCTS)~\cite{MCTSSurvey}.
Monte Carlo tree search is a planning algorithm for sequential decision-making problems and well suited for the sorting task as we will show.
First, the algorithm can search in high-dimensional state spaces by only performing a forward search. This allows us to employ commonly available physics models to predict the outcome of pushing actions that involve complex multi-object, multi-contact dynamics.
Second, the algorithm employs an adaptive sampling strategy that focuses its search on the parts of the state space that are relevant to solving the problem.
This is particularly important as the sorting task has a large state space and modeling its multi-contact physics is computationally expensive.
Third, MCTS requires no explicit target states but instead can be applied when there is only a discriminative function to evaluate whether a state is a goal.
This is the case in the sorting task, where the goal is defined through relative positions of objects rather than absolute target positions.

The contribution of this work lies in adapting the MCTS algorithm to the sorting task and evaluating it on a variety of scenarios.
To reduce the need for long physics rollouts, we propose a heuristic reward signal that successfully guides the algorithm towards sorted states.
Furthermore, inspired by the recent success of AlphaGo~\cite{silver2016mastering}, we train a rollout policy from data to improve the algorithm's performance further.
We evaluate the approach for different numbers of objects and classes, different object shapes (convex and non-convex), and sorting in the presence of immovable obstacles.
In addition, we evaluate the approach's effectiveness under modeling inaccuracies in simulation and on a real robot.

The remainder of this paper is structured as follows.
We first formally define our sorting task in \secref{sec:sorting},
before discussing related work in more detail in \secref{sec:relatedwork}.
Thereafter, we provide background information on MCTS in \secref{sec:background}
and present our adaptations in \secref{sec:method}. We present experimental
results in \secref{sec:experiments} and conclude in \secref{sec:conclusion}.

%% file: problem-def.tex

\section{PROBLEM DEFINITION}
\label{sec:sorting}
In the planar push sorting problem (PPSP), a robot $\mathcal{R}$ is tasked to sort a set of movable objects $\mathcal{M}$ in a bounded workspace according to given class membership, see \figref{fig:problem}.
The workspace is planar and all objects are assumed to be rigid.
Accordingly, the state spaces are $\mathcal{X}_i \subset \textit{SE}(2)$ for $i~\in~\{\mathcal{R}\}~\cup~\mathcal{M}$,
and the composite state space of the world is $\mathcal{X} = \mathcal{X}_1 \times \ldots \mathcal{X}_{|\mathcal{M}|} \times \mathcal{X}_\mathcal{R}$.
The workspace may contain immovable obstacles $\mathcal{O}$ that needs to be avoided.
If there are no two objects intersecting and no collisions with obstacles, we refer to states $x \in \mathcal{X}$ to be valid.

Each movable object belongs to exactly one class $c \in \mathcal{C}$.
These classes are user-defined and can be based on, for example, shared physical properties
(e.g., color, shape, size) or a common functional purpose of the objects.
The task of the sorting problem is to rearrange the objects
into a \textit{sorted} valid state according to their class membership.
A \textit{sorted} state is a state where the objects of each class form disjoint clusters, see Fig.~\ref{fig:problem}.
More formally, let $\text{CH}(c_i, x)~\subset~\mathbb{R}^2$ denote the smallest convex set that contains
all objects of class $c_i \in \mathcal{C}$ in state $x \in \mathcal{X}$.
Furthermore, let the function
\begin{equation}
    d_{c}(A, B) =
    \begin{cases}
        \min_{a \in A, b \in B} \|a - b\|_2 & \text{if~} A \cap B = \emptyset \\
        0 & \text{else.}
    \end{cases}
    \label{eq:convex_distance}
\end{equation}
denote the smallest pairwise distance between elements of two sets, $A, B \subset \mathbb{R}^2$.
We define a state $x \in \mathcal{X}$ to be \textit{sorted},
if all classes have at least a distance $\epsilon > 0$
from each other and the obstacles:
\begin{equation}
    \begin{split}
    \texttt{Sorted}(x) :\iff &
    \min_{\substack{i, j \in \mathcal{C} \\ i \neq j}} d_{c}(\text{CH}(i, x), \text{CH}(j, x)) > \epsilon \\
    \wedge & \min_{i \in \mathcal{C}}d_c(\text{CH}(i, x), \mathcal{O}) > \epsilon.
    \end{split}
\label{eq:discriminator}
\end{equation}

Let $\mathcal{A}$ be the set of planar motions that $\mathcal{R}$ is able to execute.
We limit $\mathcal{A}$ to motions that are sufficiently slow, so that pushing dynamics can be assumed to be quasistatic~\cite{Lynch1996}.
Given an initial valid state $x_0 \in \mathcal{X}$, the problem of planar push sorting is then to compute and execute a series of actions $a \in \mathcal{A}$ that transfer the system from state $x_0$ to any valid sorted state $x_g \in \mathcal{X}$.
In this process, all intermediate states $x_i$ have to be valid, i.e., not colliding with any immovable obstacles or be out of bounds.

We consider this problem in scenarios where objects are initially densely packed.
In addition, objects of the same class eventually need to be pushed into the same region.
This renders the ability to purposefully push multiple objects simultaneously essential to efficiently solve this task.

%% file: relatedwork.tex
\section{RELATED WORK}
\label{sec:relatedwork}

\subsection{Non-prehensile Rearrangement}
Non-prehensile rearrangement covers a variety of different tasks. We distinguish between navigation or manipulation among movable obstacles and large-scale rearrangement tasks.
In navigation or manipulation among movable obstacles, the priority is to navigate the robot or transport individual objects in the presence of clutter.
In other words, the goal is expressed with respect to a few individual objects or the robot, while the remaining objects may be placed anywhere.
This category includes repositioning tasks~\cite{King2015,King2016,King2017,Haustein2015,Haustein2018,Bejjani2018,Pinto2018},
reaching for an object within clutter~\cite{Muhayyuddin2018,Elliott2016,Kitaev2015,Agboh2018,Laskey2016},
as well as singulating or separating individual objects~\cite{Chang2012,Hermans2012,Eitel2017,Danielczuk2018}.
By large-scale rearrangement, we refer to problems where the goal is expressed in terms of many objects and all final poses are relevant to the task.
Our sorting task is such a problem, and to the best of our knowledge, only Huang et al.~\cite{Huang2019} have previously addressed such problems in combination with multi-object pushing.

Huang et al. found that iterative local search (ILS) equipped with strong heuristics and an $\epsilon$-greedy rollout policy succeeds at solving various table-top rearrangement tasks, including a sorting task of up to $100$ cubes.
The addressed sorting problem, however, differs from ours in two key aspects.
First, for the sorting goal, explicit target locations for each class are provided as input. This allows to derive a heuristic for action sampling, as it can easily be determined which objects are misplaced and where they should be pushed to.
In our problem, in contrast, no explicit target locations are provided, and instead, the planner needs to select suitable locations to achieve a sorted state itself.
Second, the problem specifically addresses table-top sorting with a manipulator that is capable of moving the pusher in and out of the pushing plane at any location.
In our problem, the pusher's motion is constrained to the pushing plane, requiring it to circumnavigate objects and obstacles.
Designing a strong rollout policy for such navigation tasks---as needed for ILS---is non-trivial. MCTS, in contrast, does not require similar heuristics and succeeds even with a random rollout policy, as our experiments will demonstrate.

The additional challenges of our sorting problem are useful to consider for two reasons.
First, relieving the user from providing a sorted target state as input makes the algorithm easier to use.
Second, constraining the pusher's motion to the plane is more general.
It applies to mobile robots, as well as to manipulators that have few degrees of freedom, such as Delta robots.
In addition, although not explicitly studied in this work, our problem formulation resembles a sorting task in constrained spaces such as shelves and may provide relevant insights for future work in this direction.

\subsection{Monte Carlo Tree Search for Rearrangement Planning}
Monte Carlo tree search has recently been applied to rearrangement planning problems using pick-and-place.
Zagoruyko et al.~\cite{Zagoruyko2019} use MCTS to rearrange up to $9$ objects to user-given target pose.
The authors also train a rollout policy from solutions produced by MCTS that makes the algorithm efficient enough to replan online and thus compensate for disturbances during the execution.

King et al.~\cite{king2017unobservable} proposed to apply MCTS for pushing a single object among movable obstacles under uncertainty.
This approach focuses on computing the most robust sequence of pushing actions by planning on belief space.
Here, the adaptive sampling of MCTS makes this tractable by focusing the limited computational budget for constructing a state belief model only on the most promising trajectories.

In contrast, we employ MCTS on a large-scale non-prehensile sorting task and address uncertainty only indirectly through replanning.
For this, we equip the algorithm with a heuristic reward signal that allows us to limit the computationally expensive rollouts, making the algorithm efficient enough for replanning after each push.

%% file: method.tex

\section{BACKGROUND}
\label{sec:background}

\begin{figure}[t]
\centering
\includegraphics[width = 1.0 \linewidth]{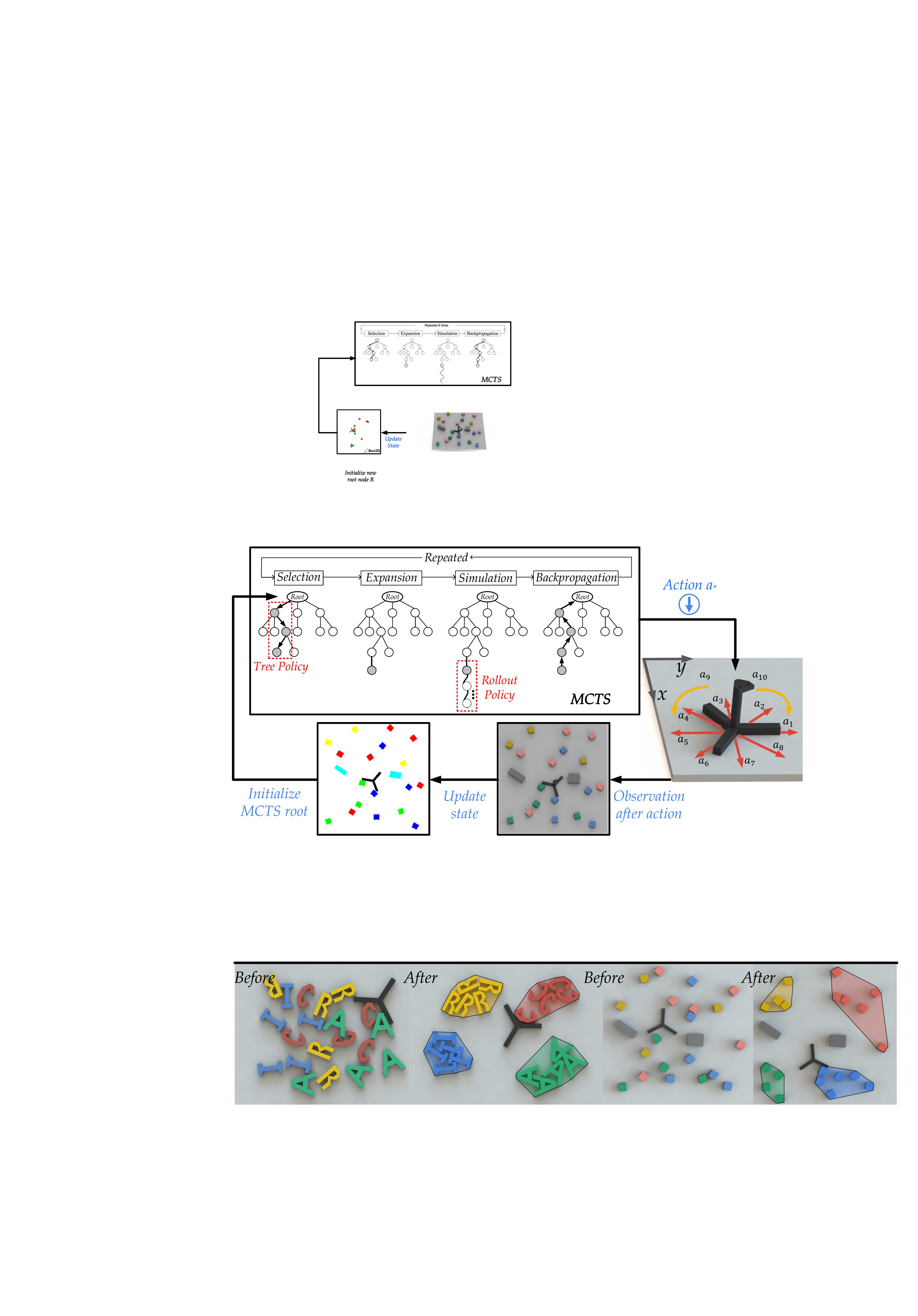}
\caption{The sorting planner's scheme for real-world execution.}
\label{fig:planner}
\end{figure}

In this section, we describe the basic form of Monte Carlo tree search (MCTS)~\cite{MCTSSurvey}, which we use in our planar push sorting planner described in~\secref{sec:method}.
MCTS is used for sequential decision-making problems with state space $\mathcal{X}$, action space $\mathcal{A}$, reward signal $g$, and transition model $\Gamma$.
Given a current state $x_t \in \mathcal{X}$ the algorithm estimates the state-action value function
$Q(x_t, .)$ using simulated episodes called rollouts and returns the best action.
During rollout, it uses a (often simple) rollout policy to decide on actions and simulates state transitions using $\Gamma$.
To focus on high-reward regions, MCTS builds a search tree with states as nodes and actions as edges, rooted in the current state.
With this tree, the algorithm maintains value estimates for the states that are most likely to be reached within a few steps.
For every single iteration, the algorithm executes the following steps which are also
shown in \figref{fig:planner}:

\textbf{Selection:}
Use a tree policy $\pi_\text{tree}$ to traverse from the root to a leaf node.
The tree policy exploits the state-action value estimates for the states in the tree and balances exploration and exploitation.

\textbf{Expansion:}
Expand the search tree by selecting an action and adding the reached state as a child node.

\textbf{Simulation:}
Use the rollout policy $\pi_\text{roll}$ for action selection and simulate the episode until termination according to $\Gamma$.

\textbf{Backup:}
Use the return generated by the episode to update the state-action value estimated for the traversed edges in the search tree.

Often, MCTS is set to terminate after a certain number of iterations $n_\text{max}$ or through some statistical criterion.
Nodes within the tree are first fully expanded before any of their children are expanded.
If terminal states are too many steps away for full rollout simulation to be tractable---as it is in our case---truncated rollouts are used.
Then, instead of the return of the complete episode, some heuristic estimate of the return is backed up.
Due to the policy improvement theorem, selecting the best action at the root node is at least as good as the rollout policy.
However, because the value estimates are based on long-term consequences, it is usually better.
The details of how we adapt MCTS to the planar push sorting problem are explained in the following section.

\section{PLANAR PUSH SORTING PLANNER}
\label{sec:method}

In this section, we first model the PPSP as a sequential decision-making problem for MCTS.
Then, we outline how our algorithm uses MCTS and decides on termination.
We describe how we simulate PPSP in Monte Carlo rollouts and detail the four MCTS steps mentioned in \secref{sec:background}.
Finally, we explain how we use deep learning to obtain a rollout policy from data to improve our MCTS.

\subsection{Sequential Decision-making Problem}

To model the PPSP we use the full configuration space $\mathcal{X}$ as the state space and define a robot-centric action space $\mathcal{A}$ with 10 actions as depicted in \figref{fig:planner} on the right.
The robot can translate into 8 different directions and rotate left and right in small increments.
As transition model $\Gamma$, we employ the physics simulator Box2D,\footnote{Box2D, A 2D Physics Engine for Games: \url{https://box2d.org/}} which is capable of modeling multi-object interactions.
We model the PPSP as a deterministic process and compensate for errors in the physics modeling by replanning after each push.

With our action space, it often takes up to 200 transitions until a sorted state is reached, which makes large numbers of full rollouts with the physics simulation practically intractable.
For this reason, we use truncated rollouts of length $d_\text{max}$.
However, this means that we cannot use $\texttt{Sorted}(x)$ from~\eqref{eq:discriminator} as a feedback signal since most rollouts do not reach a sorted state.
Therefore, we define a different reward signal $g(x)$ that provides useful feedback also for
unsorted states and increases when the state becomes more sorted.

\textbf{Reward signal:}
We construct the reward signal for a state $x$ from four components:
A measure how compact a class $c_i$ is, $E_i^\text{self}(x)$,
how far away class centers are from each other, $E_{ij}^\text{other}(x)$,
how far away class centers are from obstacles, $E_{i}^\text{obst}(x)$,
and the distance of the two closest class centers, $d_\text{cent}(x)$.
Concretely, we define:
\begin{align}
&E_{i}^\text{self}(x)
=
\frac{1}{\left| c_i \right|}\sum\limits_{m \in c_i} \ln(p_i(x_{m}))
\label{eq:self}
\\
&E_{ij}^\text{other}(x)
=
\ln(1-p_i(\mu_{j}(x)))
\label{eq:other}
\\
&E_{i}^\text{obst}(x)
=
\sum_{o\in\mathcal{O}}\ln(1-p_i(\mu_{o}))
\label{eq:obst}
\\
&d_\text{cent}(x)
=
\min\limits_{i, j \in \mathcal{C}, i \neq j} \|\mu_{i}(x) - \mu_{j}(x)\|_2,
\end{align}
where $\mu_{i}(x)$ denotes the mean position of all objects in class $c_i$,
$\mu_{o}$ the centroid of obstacle $o$,
$x_m$ the position of object $m$, and $p_i(x_{m})= e^{-\lambda({\| x_{m} - \mu_{i} \|}_2^2)}$
a Gaussian with variance $\lambda$ centered at the mean position of class $c_i$.
The term $E_i^\text{self}(x)$ increases as objects of class $c_i$ are more compact.
The term $E_{ij}^\text{other}(x)$ increases as the centers of class $c_i, c_j$ are moved apart.
Similarly, the term $E_i^\text{obst}(x)$ increases as the center of class $c_i$ is moved away from obstacles.

We combine these terms together to form the reward signal
\begin{equation}
g(x)= \frac{\sum\limits_{i=1}^{\left| \mathcal{C} \right|} \left(E_i^\text{self}(x) + \sum\limits_{j=1}^{i-1}E_{ij}^\text{other}(x) + E_i^\text{obst}(x)\right)}{d_\text{cent}(x)}.
\label{eq:state}
\end{equation}
It is $g(x) < 0$ for all states, but it approaches $0$ as members of the same class get closer and members of different classes separate.
It eventually reaches higher values for sorted states (as defined in \eqref{eq:discriminator}) than for unsorted states.
Hence, by maximizing $g(x)$, our planner will gradually aggregate objects of the same class and separate those from different classes.

\subsection{Sorting Planner Outline}

Starting with an unsorted state $x$, our sorting algorithm repeatedly runs MCTS to obtain the best action for this state, executes the action and observes the result.
This process is illustrated in \figref{fig:planner} and \algref{alg:planner}.
In case we reach a sorted state, the sorting algorithm terminates.
Additionally, we apply two strategies to detect whether the planner fails in the sorting process.
First, it counts the number of subsequent actions where the robot has not pushed a single object and terminates if this number exceeds a conservatively chosen threshold.
Second, it compares the best reward observed for any visited state during any of the rollouts in MCTS, $\hat g$, with the reward of the current state $g(x)$.
If the relative difference $\frac{\hat{g} - g(x)}{|g(x)|}$ is smaller than a small threshold $\nu > 0$, the planner also returns failure since there is no perspective of improving the state any further.

\begin{algorithm}[t]
\caption{Planar Push Sorting Planner}
\label{alg:planner}
\footnotesize
\DontPrintSemicolon


\SetKwInOut{Symbols}{Symbols}
\SetKwInOut{OR}{or}
\SetKwFunction{TreeNode}{TreeNode}
\SetKwFunction{TreePolicy}{TreePolicy}
\SetKwFunction{Expand}{Expand}
\SetKwFunction{BackPropagate}{BackPropagate}
\SetKwFunction{GetNodeState}{GetState}
\SetKwFunction{BestAction}{BestAction}
\SetKwFunction{Rollout}{Simulate}
\SetKwFunction{Observe}{Observe-State}
\SetKwFunction{IsGoal}{Sorted}
\SetKwFunction{MCTS}{MCTS}
\SetKwFunction{Execute}{Execute}
\SetKwFunction{Trapped}{IsTrapped}
\SetKwProg{function}{function}{}{}

$x \gets$ Observe the state\;
\While{\textbf{not} \IsGoal{$x$}  } {
$a, \hat{g} \gets$ \MCTS{$x$}\;
\lIf{\Trapped{$x, \hat{g}$}} {Terminate with \emph{Failure}}
\Execute{$a$}\;
$x \gets$ Observe the state\;
}
Terminate with \emph{Success}\;

\hrule

\function{\MCTS{$x_0$}}{
Initialize the search tree with root $x_0$ \\
$\hat{g} \gets g(x_0)$; 
$i \gets 0$\;
\For{$i < n_\text{max}$}{
Select the root node\;
\While{\textnormal{Node is fully expanded}}{
Select child node according to $\pi_\text{tree}$\;
}
Expand the current (leaf) node using $\pi_\text{roll}^0$\;
$g_\text{max} \gets$ Truncated simulation rollout using $\pi_\text{roll}$ \;
Backup with return $g_\text{max}$\;
$\hat{g} \gets \max(g_\text{max}, \hat{g})$\;
}
}
$a^* \gets$ Greedy action selection for $x_0$\;
\Return{$a^*, \hat{g}$}\;
\end{algorithm}

%% file: method-mcts.tex
\subsection{Monte Carlo Tree Search Implementation}
\label{sec:mcts}

We adapt the MCTS algorithm to our deterministic modeling and reward signal $g(x)$.
The pseudo-code is shown in \algref{alg:planner}.
In the search tree of MCTS, every node corresponds to a state $x \in \mathcal{X}$.
Accordingly, the root node corresponds to the current state $x_{t}$.
For each node $s$, the search tree stores the visitation count $N(s)$,
and estimates of an upper and lower bound of the reachable reward from its state,
$\hat{V}_\text{upper}(x)$ and $\hat{V}_\text{lower}(x)$, respectively.

\textbf{Selection} and \textbf{Expansion} are executed as described in
\secref{sec:background} and we use the tree policy $\pi_\text{tree}$ as
defined below.

\textbf{Simulation:}
We use the simulator $\Gamma$ to predict the effect a robot action has as it pushes through objects.
The actions are selected either from a random rollout policy $\pi_\text{roll}$ or from a learned rollout policy which is detailed below.
The rollout is truncated after $d_\text{max}$ steps to save computation.
Since the rollout policy is sub-optimal---e.g., it may select actions which worsen previous gains---we return the maximal reward signal from the rollout $g_\text{max}$ instead of the reward signal at the truncation state as a heuristic estimate of the episode's return.

\textbf{Backup:}
When a rollout is finished, we increment the visitation counter $N(s)$ and update the bound estimates for each traversed node $s$ in the search tree, %
\begin{align}
\hat V_\text{upper}(x) &\gets g_\text{max}, \qquad \textrm{if}\, g_\text{max} > \hat V_\text{upper}(x)
\\
\hat V_\text{lower}(x) &\gets g_\text{max}, \qquad \textrm{if}\, g_\text{max} < \hat V_\text{lower}(x)
,
\end{align}
where $x$ is the state for node $s$.


\textbf{Tree policy:}
During selection, we use a tree policy $\pi_\text{tree}$ that balances between exploration and exploitation and additionally considers the visitation count.
For this, we estimate the state-action value $Q(x, .)$ at a node as:
\begin{equation}
Q(x, a_i)
=
\frac{
\hat V_\text{upper}(x_i)- \hat V_\text{lower}(x)}{
\hat V_\text{upper}(x)-\hat V_\text{lower}(x)}
+
C\sqrt{\frac{2\ln N(s)}{N(s_i)}}
\label{eq:ucb_max}
\end{equation}
where $x_i$ is the state of the child node corresponding to action $a_i \in \mathcal{A}$.
The formula is derived from UCB1~\cite{auer2002finite}, but different from the standard UCT approach~\cite{MCTSSurvey} in that we normalize rewards and use the maximum reward rather than the average.
We choose this modification, as the maximum reward is a good estimator for Monte Carlo rollouts under a deterministic model (i.e., simulator).
The exploration term $C$ is set to $\frac{1}{\sqrt{2}}$.

\textbf{Rollout policy:}
While a completely random rollout policy guarantees probabilistic completeness, it is much more sample effective to select actions in an informed way during simulation.
For this reason, we learn a rollout policy from the successful sorting experience of a random rollout policy $\pi_\text{roll}^0$ as detailed in Sec.~\ref{sec:policy} below.

%% file: method-policy.tex

\subsection{Learning the Rollout Policy}
\label{sec:policy}
\label{sec:representation}
\label{sec:network}
\label{sec:collection}

The rollout policy $\pi_\text{roll}$ has to map states $x \in \mathcal{X}$ to actions $a \in \mathcal{A}$ and, after the physical simulation, presents the second computational bottleneck during Monte Carlo rollout.
In our case, the state space can also be different for different numbers of objects and obstacles.
For this reason, we encode the positions of objects, obstacles and the robot in the state $x$ as a 2D color image $\mathcal{I}(x) \in [0, 1]^{256 \times 256 \times 3}$, which shows the footprint of each object colored by class.
This high-dimensional representation $\mathcal{I}(x)$ requires a policy network with many parameters and thus would require large amounts of labeled training data.
To alleviate this data acquisition problem, we first learn a lower-dimensional embedding of the image space $f\colon \mathcal{I}(x)~\mapsto~\tilde x~\in~[0, 1]^{32 \times 32 \times 3}$ from unlabeled image data.
For the rollout policy, we then learn a mapping from this lower-dimensional space to probability distributions over actions, $P$.

\begin{figure}[]
\centering
\includegraphics[width = 1 \columnwidth]{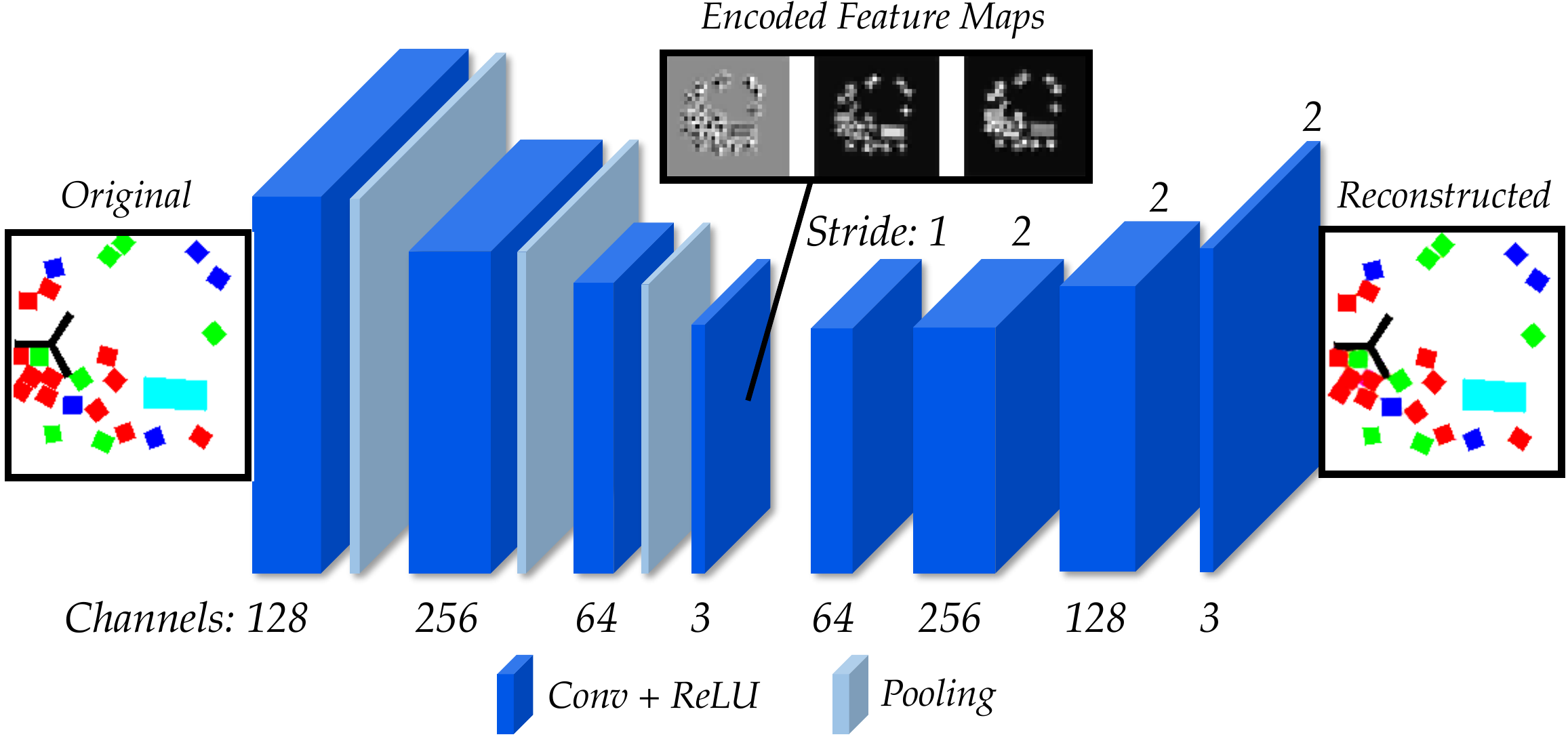}
\caption{Architecture of the autoencoder.}
\label{fig:encoder}

\centering
\includegraphics[width = 1 \columnwidth]{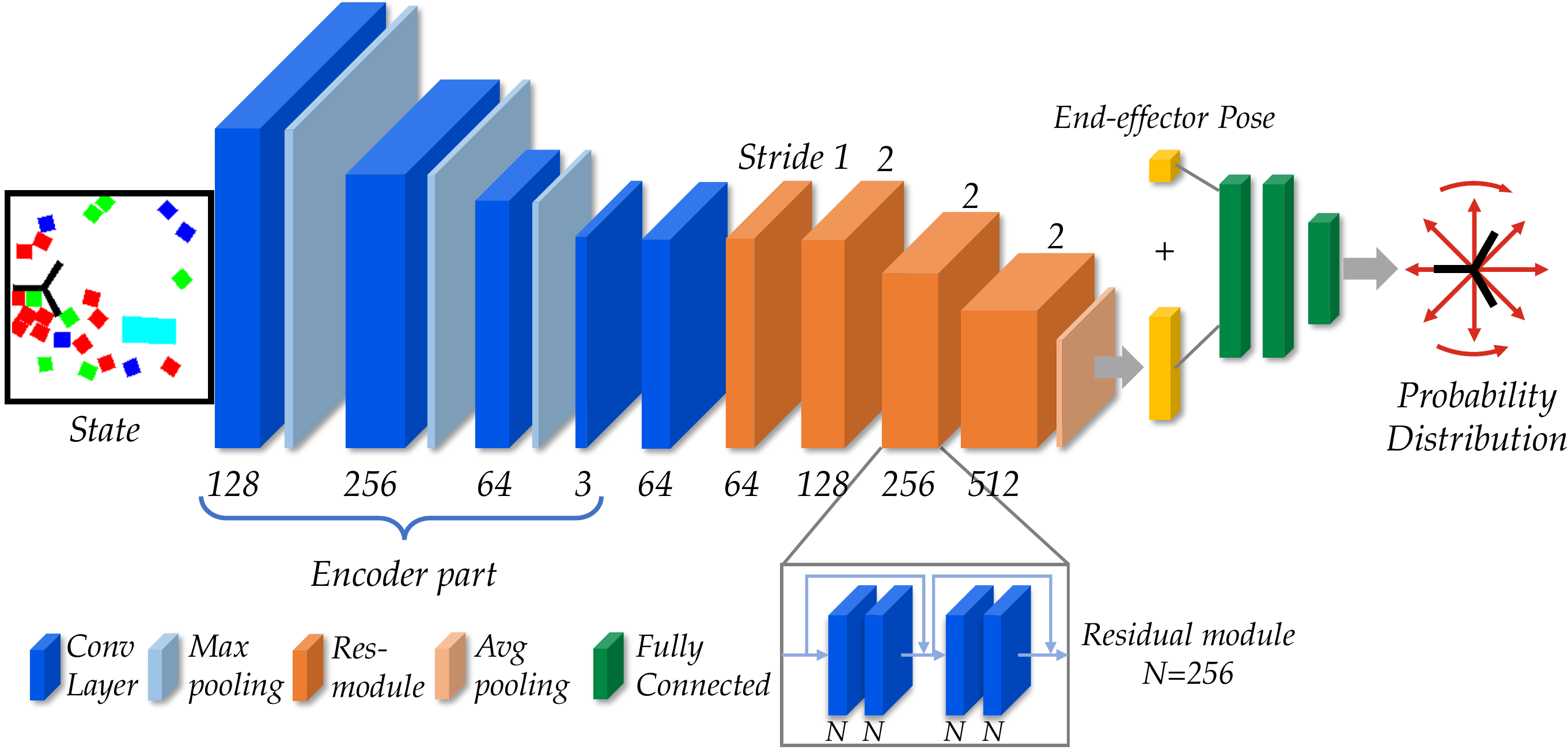}
\caption{Architecture of the policy network.}
\label{fig:policy_net}
\end{figure}

\textbf{State representation learning:}
We model the embedding function $f$ as the encoder module in a convolutional autoencoder as seen in Fig~\ref{fig:encoder}.
The training data is generated from our simulator, consisting of $120$k images with each image showing $20$-$30$ square objects.
The training objective is the mean squared reconstruction loss.

\textbf{Rollout policy learning:} We model the rollout policy
$\pi_\text{roll} \colon (x_\mathcal{R}, \tilde x) \mapsto (P(a_1), \dots, P(a_{10}))$
with the deep convolutional architecture ResNet-18, which is
reported to be easier to optimize and resilient against overfitting
\cite{He2015}, as depicted in \figref{fig:policy_net}. The ResNet structure
was initially designed for image classification, while here it is employed for
mapping the state features to action labels. To produce the labeled training data, we run
our planar push sorting planner (see \algref{alg:planner}) as described
above with a fully random rollout policy $\pi_\text{roll}^0$ for sorting
$16$, $20$ and $24$ cubes with one static obstacle randomly placed in the
scenes. We record the observed transitions, $16.5$k in total, for each solved sorting problem as
tuples $\left(\mathcal{I}(x), x_{\mathcal{R}}, Q(\mathcal{A}) \right)$, where
$Q(\mathcal{A})$ are the exploitation terms of the state-action value
estimates for $x$ in \eqref{eq:ucb_max}.

We train the policy model with cross-entropy loss while keeping the encoder parameters fixed.
The training target is the probability distribution over actions which arises from normalizing the state-action values, i.e. $P(a_i)~=~Q(a_i) /\sum \limits_{a \in \mathcal{A}}Q(a)$.

%% file: experiments.tex

\section{EXPERIMENTS}
\label{sec:experiments}

In this section, we evaluate the performance of the proposed algorithm.
First, in \secref{subsec:baselines}, we conduct experiments to motivate the choice of MCTS and compare the algorithm's performance against several baselines in simulation.
Second, in \secref{subsec:diff_tasks}, we quantitatively evaluate how the algorithm performs in different sorting tasks, including 1) different number of objects and classes, 2) non-convex objects, and 3) convex objects in the presence of immovable obstacles.
In addition, in the latter case, we evaluate how the learned rollout policy improves performance.
Third, to investigate how our approach performs under modeling errors, we provide quantitative results under different degrees of simulated noise and on a real robot in \secref{subsec:uncertainty}.
Lastly, we report the runtime of our implementation in \secref{subsec:runtime}.

\begin{table*}[t]
\begin{center}
  \begin{tabular}{cc|ccccccc}
  \toprule
  \multicolumn{2}{c}{Methods}&
  \textbf{Greedy-one-step} & \textbf{Greedy-rollout} & \textbf{MCTS-no-rollout} & \textbf{ILS-3}  & \textbf{ILS-6} & \textbf{Ours-avg} & \textbf{Ours}           \\
  \hline
  \multirow{4}{*}{2 classes}
  & \multirow{2}{*}{20 objs} & \textbf{27\%}   & \textbf{84\%}  & \textbf{87\%}   & \textbf{51\%}  & \textbf{76\%}  & \textbf{100\%} & \textbf{100\%} \\
  &                          & $154.0\pm10.7$  & $41.4\pm1.3$   & $39.8\pm1.6$    & $178.2\pm9.6$  & $218.9\pm14.9$ & $46.4\pm2.0$   & $36.1\pm1.3$   \\ \cline{3-9}
  & \multirow{2}{*}{30 objs} & \textbf{10\%}   & \textbf{41\%}  & \textbf{66\%}   & \textbf{22\%}  & \textbf{42\%}  & \textbf{97\%}  & \textbf{96\%}  \\
  &                          & $231.2\pm37.5$  & $64.4\pm3.3$   & $65.7\pm2.4$    & $294.4\pm17.4$ & $368.0\pm25.6$ & $93.5\pm3.5$   & $77.5\pm3.6$   \\ \hline
  \multirow{4}{*}{3 classes}
  & \multirow{2}{*}{20 objs} & \textbf{14\%}   & \textbf{49\%}  & \textbf{54\%}   & \textbf{40\%}  & \textbf{62\%}  & \textbf{96\%}  & \textbf{98\%}  \\
  &                          & $170.6\pm20.7$  & $61.3\pm2.3$   & $56.5\pm2.0$    & $232.2\pm17.5$ & $287.5\pm14.6$ & $81.0\pm3.1$   & $66.6\pm2.3$   \\ \cline{3-9}
  & \multirow{2}{*}{30 objs} & \textbf{3\%}    & \textbf{17\%}  & \textbf{25\%}   & \textbf{10\%}  & \textbf{19\%}  & \textbf{88\%}  & \textbf{91\%}  \\
  &                          & $291.3\pm27.5$  & $98.5\pm5.1$   & $99.2\pm4.7$    & $421.6\pm30.5$ & $535.0\pm26.9$ & $152.3\pm5.6$  & $131.5\pm5.1$  \\ \hline
  \multirow{4}{*}{4 classes}
  & \multirow{2}{*}{20 objs} & \textbf{6\%}    & \textbf{36\%}  & \textbf{43\%}   & \textbf{28\%}  & \textbf{57\%}  & \textbf{95\%}  & \textbf{97\%}  \\
  &                          & $176.7\pm15.1$  & $74.4\pm2.3$   & $71.6\pm2.8$    & $262.1\pm14.4$ & $329.3\pm16.3$ & $94.7\pm2.7$   & $80.1\pm2.3$   \\ \cline{3-9}
  & \multirow{2}{*}{30 objs} & \textbf{0\%}    & \textbf{2\%}   & \textbf{12\%}   & \textbf{4\%}   & \textbf{8\%}   & \textbf{85\%}  & \textbf{89\%}  \\
  &                          & N. A.           & $113.0\pm13.4$ & $108.1\pm4.2$   & $525.7\pm10.3$  & $726.5\pm36.0$ & $182.6\pm5.00$ & $162.6\pm4.6$  \\
  \hline
  \multicolumn{2}{c}{Average Success}&
                             \textbf{10.0\%} & \textbf{38.2\%} & \textbf{47.8\%} & \textbf{25.8\%} & \textbf{44.0\%} & \textbf{93.5\%} & \textbf{95.2\%} \\
  \bottomrule
  \end{tabular}
\end{center}
\caption{Quantitative results of comparing different algorithms on sorting 20 and 30 randomly placed cubes assigned to 2, 3 or 4 classes.}
\vspace{-0.3cm}
\label{table:baselines}
\end{table*}

\subsection{Experimental Setup}
We use three types of objects in our evaluation:
cubes of size $2.5cm \times 2.5cm$,
U-shaped non-convex objects that can surround a cube,
and randomly generated rectangular obstacles with an area no larger than twice the area of a cube.
Unless stated otherwise, all evaluations are run with the following parameters:
$\epsilon=0.05m$, $\nu=0.05$ and $d_\text{max}=3$.
All objects are placed randomly in a $50cm \times 50cm$ workspace.
The pusher's action space is set to $5cm$ translations and rotations of $45^{\circ}$.
We set the maximal number of actions the robot is allowed to execute without contacting any object to $15$.
Finally, all results are reported by the overall success rate (bold) from running 100 trials, together with the mean and standard error of the step numbers from successful runs.

\subsection{Baseline Comparison}
\label{subsec:baselines}
To motivate MCTS, we compare the following algorithms:

\textbf{Greedy-policy:} Execute the action with the maximal probability $P(a_i)$ output by the learned policy.

\textbf{Greedy-one-step:} In state $x$, execute the action $a \in \mathcal{A}$
that achieves maximal reward $g(\Gamma(x, a))$ by one-step look-ahead and use random selection for tie breaks.

\textbf{Greedy-rollout:} Run $n=500$ rollouts of depth $d_{max} = 3$ using the random rollout policy.
Execute the first action of the rollout that reaches the maximal reward $g(x)$ and use random selection for tie breaks.

\textbf{MCTS-no-rollout:}
Perform tree search as in \algref{alg:planner} but obtain rewards only from the leaf nodes without rollouts.

\textbf{Iterated Local Search (ILS):}
The algorithm as described in \cite{Huang2019} adapted to our problem.
For a fair comparison, the algorithm operates on the same discrete action space $\mathcal{A}$ and is only equipped with a random rollout policy, i.e.\ no problem-specific heuristics.
In contrast to MCTS, the ILS algorithm is designed to construct a long pushing trajectory that eventually reduces the distance to a goal state.
In our case, due to the random rollout policy, the trajectories produced by ILS vary greatly.
This poses a problem if we replan after each action.
The robot may frequently change direction, which results in little to no progress in the sorting task.
To alleviate this, we employ the following meta-algorithm for closed-loop execution. In each step, we execute ILS to generate a new trajectory and compare it to the trajectory planned in the previous step.
Only if the new trajectory is predicted to lead to a sorted or a state with greater reward $g(x)$, do we switch to the new trajectory.
Otherwise, we keep executing the old trajectory. We evaluate ILS algorithm for $n=500$ iterations of local search of depths $d_{max}=3, 6$.

\textbf{Ours-avg:} MCTS as shown in \algref{alg:planner} but with averaged rewards in \eqref{eq:ucb_max}.
The algorithm is run for $n_\text{max} = 500$ iterations.

\textbf{Ours:} MCTS as presented in \secref{sec:method}.
The algorithm is run for $n_\text{max} = 500$ iterations.

We query all algorithms to solve six cube-sorting tasks with different numbers and classes in simulation. The results are shown in \tableref{table:baselines},
in which Greedy-policy is not included as it fails in all the problems.
We observe that the Greedy-one-step performs poorly, indicating that greedily maximizing $g(x)$ is insufficient to solve the sorting tasks.
In particular, the lack of guidance on the robot motion makes this simple baseline struggle.
The Greedy-rollout achieves a good success rate on the easiest problem but rapidly declines as the problem becomes more complex.
This demonstrates the benefits of selecting actions based on long-horizon rollouts.
The same observation can be made from the performance of the MCTS-no-rollout.
While it can improve over the performance of the Greedy-rollout, it is clearly outperformed by MCTS with rollouts (Ours-avg, Ours).

While the ILS algorithm can also solve some instances of our sorting problem, it performs poorly.
This comes at no surprise, as ILS is not designed for rollout policies that are as uninformed as the random policy.
The tree search in MCTS, in contrast, improves upon its uninformed rollout policy and is capable of solving more difficult tasks at high success rates and with much fewer actions than the baselines.
Lastly, while using different state-action functions has a similar success rate,
we observe that using the maximum reward as in~\eqref{eq:ucb_max} leads to on average fewer number of required actions.

\subsection{Different Sorting Tasks}
\label{subsec:diff_tasks}
Next, we evaluate the MCTS algorithm on more challenging problems. In these experiments, we modify the algorithm
to enhance its performance while keeping runtime low.
Specifically, we introduce two additional parameters $n_\text{min}$ and $\nu_t$ to dynamically adjust the maximum number of iterations.
MCTS runs for at least $n_{min}$ iterations in each step of the sorting algorithm.
If the best encountered reward value $\hat g$ in those rollouts does not sufficiently improve over the current state's reward,
$\frac{\hat g - g(x)}{|g(x)|} < \nu_t$, the algorithm runs MCTS for additional $n_\text{min}$ iterations.
This is continued until either a sufficiently good $\hat g$ has been observed or $n_\text{max}$ iterations have been reached.
Unless stated otherwise, we set $n_\text{min}=500$, $n_\text{max}=1500$ and $\nu_t=0.2$.

\begin{table}[htb]
\begin{center}
\setlength{\tabcolsep}{0.75mm}{
  \begin{tabular}{@{}cccccc@{}}
  \toprule
  Settings & $20$ objs & $25$ objs & $30$ objs & $35$ objs  & $40$ objs                \\
  \hline
                            & \textbf{100\%} & \textbf{99\%} & \textbf{97\%} & \textbf{89\%} & \textbf{84\%}   \\
  \multirow{-2}{*}{$2$ cls} & $34.6\pm1.8$   & $56.5\pm2.7$  & $72.2\pm3.0$  & $98.0\pm4.9$  & $121.0\pm6.0$   \\
                            & \textbf{99\%}  & \textbf{95\%} & \textbf{94\%} & \textbf{87\%} & \textbf{77\%}   \\
  \multirow{-2}{*}{$3$ cls} & $65.5\pm2.3$   & $85.4\pm3.7$  & $113.4\pm4.6$ & $146.5\pm5.9$ & $185.6\pm9.3$   \\
                            & \textbf{99\%}  & \textbf{94\%} & \textbf{90\%} & \textbf{78\%} & \textbf{71\%}   \\
  \multirow{-2}{*}{$4$ cls} & $78.5\pm2.7$   & $109.9\pm3.9$ & $159.6\pm6.2$ & $202.6\pm8.4$ & $234.2\pm11.0$  \\
  \hline
                            & \textbf{99.3\%} & \textbf{96.0\%}  & \textbf{93.7\%} & \textbf{84.7\%} & \textbf{77.3\%} \\
  \multirow{-2}{*}{Avg}     & $59.5$           & $84.0$          & $115.1$         & $149.0$         & $180.3$         \\
  \bottomrule
  \end{tabular}
}
\end{center}
\caption{Quantitative results of sorting different number of objects and classes.}
\label{table:diff_cls_objs}
\end{table}

Similar to the experiments in \secref{subsec:baselines}, we first query the algorithm to sort more randomly placed cubes.
The results are shown in \tableref{table:diff_cls_objs}.
For the more complex test cases with up to $40$ objects, we observe success rates of more than $75\%$.
According to the increased complexity, we observe much larger numbers of required actions than in the simpler cases.

\begin{figure}[htb]
\begin{center}
  \includegraphics[width=0.9\columnwidth]{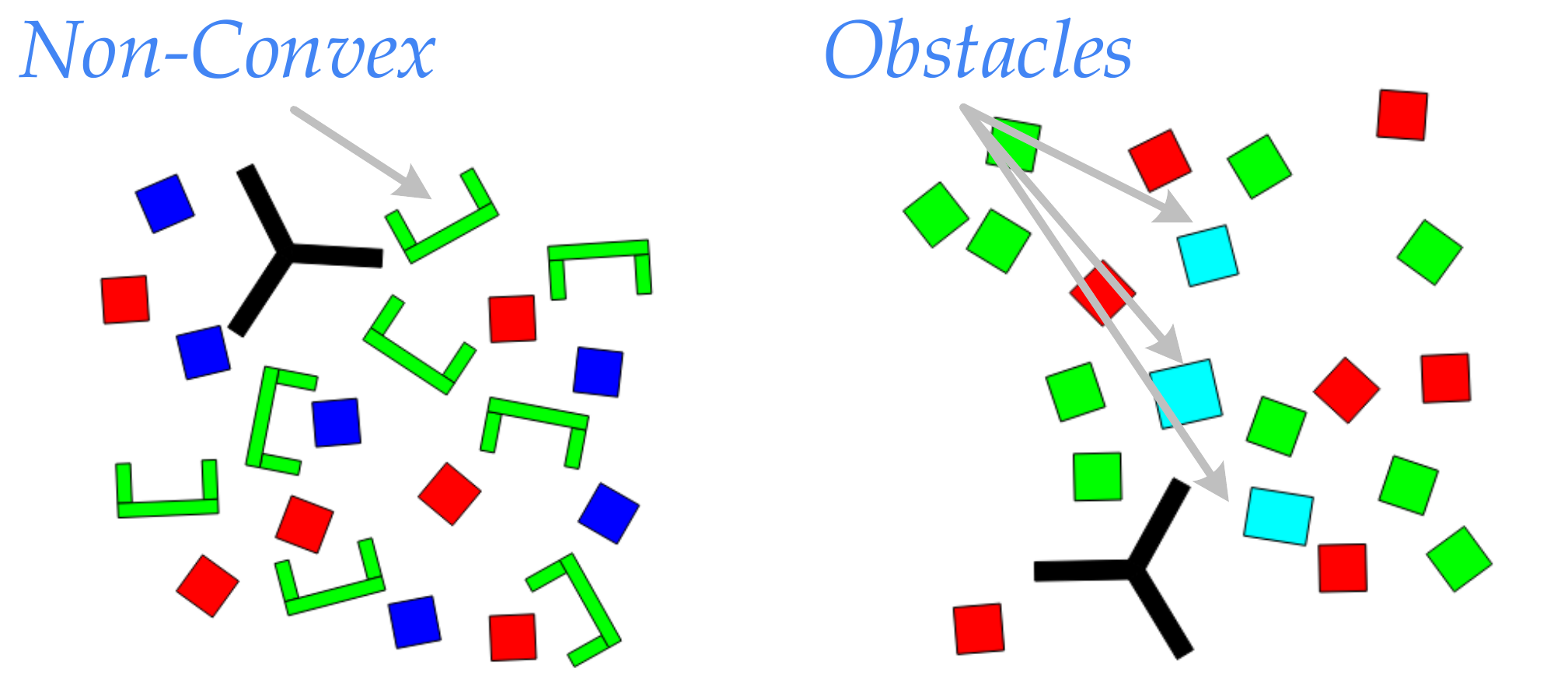}
  \setlength{\tabcolsep}{0.85mm}{
  \footnotesize
  \begin{tabular}{@{}ccccccc@{}}
  \toprule
  Non-convex
  & 10\%     & 20\%     & 30\%     & 40\%     & 50\% \\
  \midrule
  Success
  & \textbf{100\%} & \textbf{96\%} & \textbf{98\%} & \textbf{94\%} & \textbf{95\%} \\
  \# Steps
  & $64.7\pm2.9$ & $71.2\pm4.9$ & $84.1\pm3.7$ & $96.9\pm3.8$ & $95.2\pm4.3$ \\
  \bottomrule
  \end{tabular}
  }
\end{center}
\caption{(Top) example scenes with non-convex objects and obstacles. (Bottom) quantitative results of sorting $20$ objects with $10\% \sim 50\%$ ratios of non-convex objects.}
\label{fig:exp-scenes}
\end{figure}

Next, we query the algorithm to sort scenes containing cubes and non-convex U-shaped objects, see~\figref{fig:exp-scenes}.
The U-shaped objects can easily entangle, or trap a cube, which makes it hard to rearrange these objects.
The results shown in~\figref{fig:exp-scenes} indicates the algorithm can well handle this case, and the success rate is not significantly influenced.
However, the step number grows as the ratio of convex objects increases.
This is due to the fact that the robot needs to spend additional actions on disentangling objects from each other.

The third experiment is to sort 20 cubes of 2 classes in the presence of $1$, $2$ and $3$ immovable obstacles, see \figref{fig:exp-scenes}.
These are the most difficult problems, as it is hard to recover from pushing an object too close to an obstacle.
Moreover, the robot's motion is more constrained and it needs to circumnavigate obstacles.
In this experiment, the learned rollout policy and the random one are compared.
For this, we run the planner with different parameter settings
for the number of iterations $n_\text{min}$, $n_\text{max}$ and rollout depth $d_\text{max}$
($(500, 1500, 6)$, $(500, 1500, 3)$ and $(500, 500, 6)$ respectively).

\begin{figure}[htb]
\centering
  \begin{subfigure}{0.49\columnwidth}
    \centering
    \includegraphics[width=1\columnwidth]{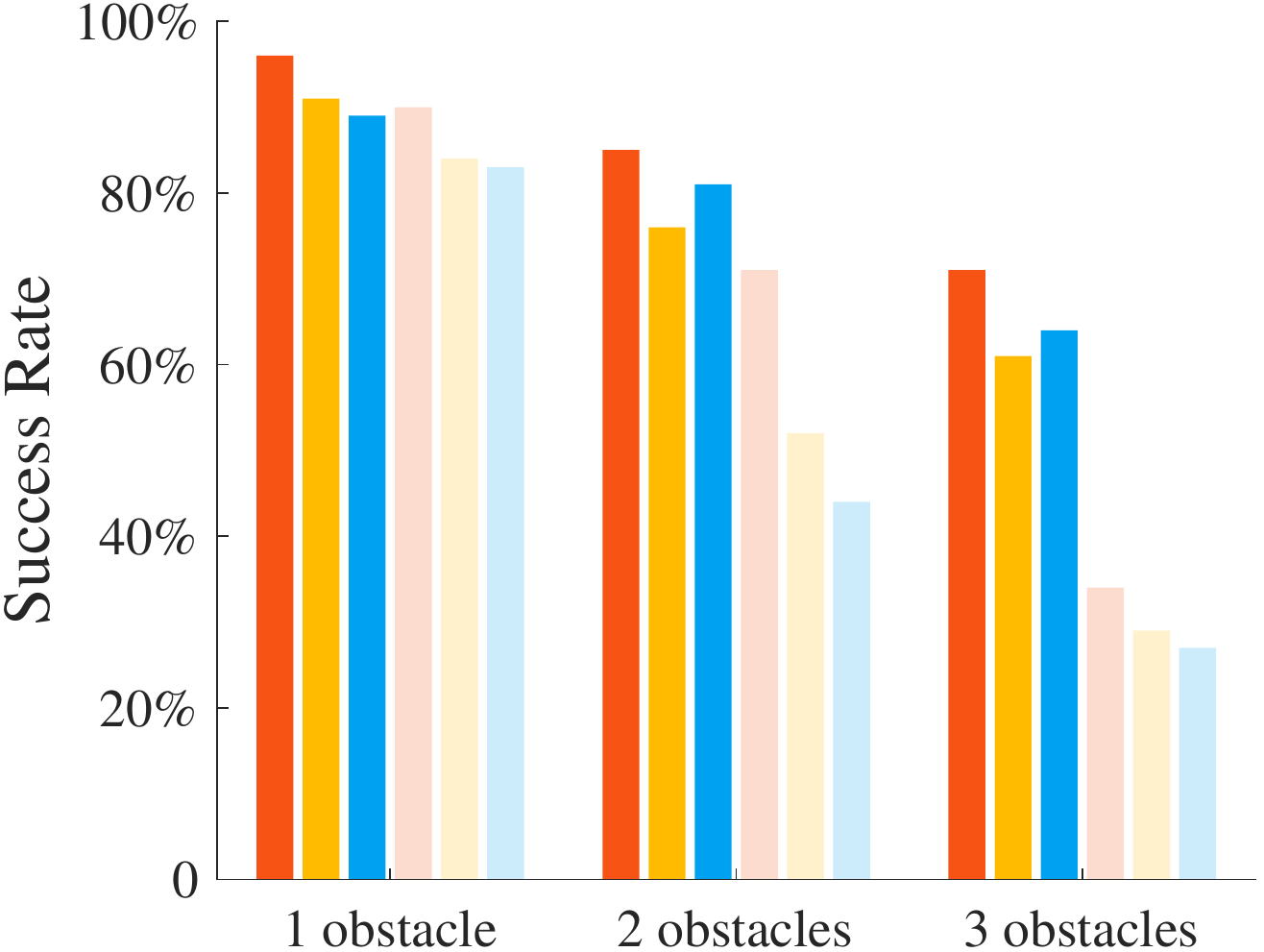}
  \label{fig:obstacles-test-success}
  \end{subfigure}
  \begin{subfigure}{0.49\columnwidth}
    \centering
    \includegraphics[width=1\columnwidth]{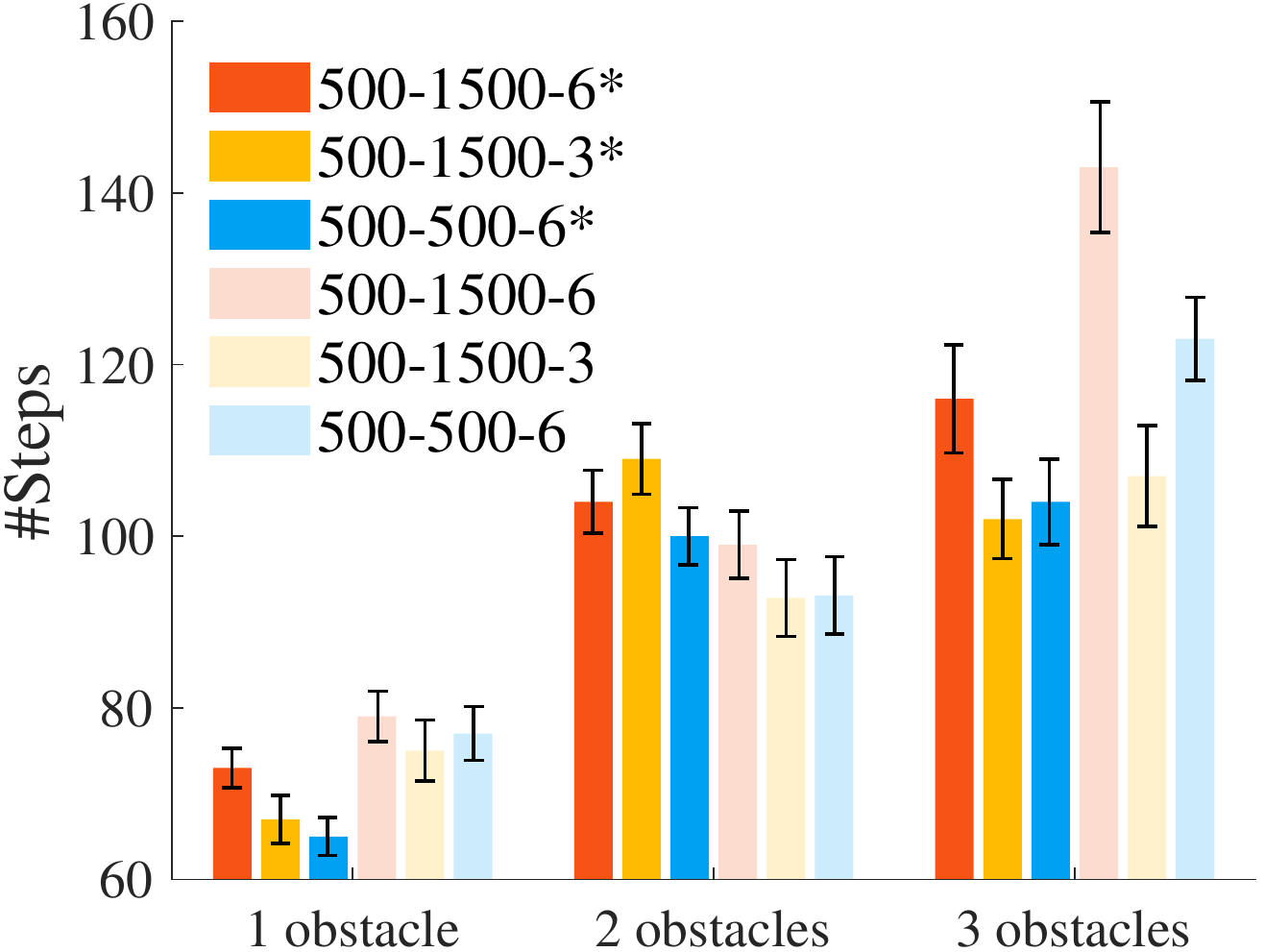}
  \label{fig:obstacles-test-steps}
  \end{subfigure}
\caption{
  Quantitative results of success rates (left) and numbers of steps (right) for sorting 20 cubes in the presence of immovable obstacles.
  Different parameter settings are presented by bars with different colors,
  where solid colors denote the use of the learned policy and transparent colors denote the use of the random policy.
}
\label{fig:plots}
\end{figure}

Recall that the learned policy is trained from sorting cubes in scenes with only one immovable obstacle.
The results shown in \figref{fig:plots} indicate that the MCTS with the learned policy successfully generalizes to problems with more obstacles.
Further, the more obstacles, the greater is the benefit of using the learned policy over the random one.
By comparison with the zero success rate of using Greedy-policy, it further validates that MCTS provides a gain over the rollout policy.
For both policies, a larger rollout depth and more iterations are unsurprisingly beneficial. Accordingly,
the dynamic adjustment of iterations leads in both cases to an improvement in success rate.
It is noteworthy that the learned policy can even achieve better results than the random one
when using fewer iterations or a shorter rollout depth.

\subsection{Evaluation Under Uncertainty}
\label{subsec:uncertainty}
In our last experiments, we evaluate the algorithm's performance under modeling error in the physical parameters.
We first perform experiments in simulation to solve different cube-sorting tasks with a fixed number of iterations $n_\text{max}=500$.
To simulate modeling error, we add Gaussian noise $\mathcal{N}(0, p\cdot{c_f})$ on the contact friction coefficient $c_f$ between objects and the ground
when simulating the execution of an action. We test with mild noise $p = 0.5$ and severe noise $p = 0.75$. The results are shown in~\tableref{table:noise}.
In most cases, the noise has little effect on the success rate. We can observe, however, an increase
in the average number of actions needed, which indicates an error-correcting behavior.

\begin{table}[t]
\begin{center}
\begin{tabular}{cc|ccc}
\toprule
\multicolumn{2}{c}{Noise Level} & No Noise & Mild $(50\%)$ & Severe $(75\%)$ \\ \hline

&                           & \textbf{100\%} & \textbf{100\%} & \textbf{99\%} \\
& \multirow{-2}{*}{20 objs} & $36.1\pm1.3$   & $36.1\pm1.2$   & $39.6\pm2.0$  \\ \cline{3-5}
&                           & \textbf{96\%}  & \textbf{97\%}  & \textbf{97\%} \\
\multirow{-4}{*}{2 classes}
& \multirow{-2}{*}{30 objs} & $77.5\pm3.6$   & $78.1\pm3.2$   & $81.8\pm3.9$  \\ \hline

&                           & \textbf{98\%}  & \textbf{96\%}  & \textbf{95\%} \\
& \multirow{-2}{*}{20 objs} & $66.6\pm2.3$   & $71.9\pm2.9$   & $83.2\pm3.7$  \\ \cline{3-5}
&                           & \textbf{91\%}  & \textbf{91\%}  & \textbf{91\%} \\
\multirow{-4}{*}{3 classes}
& \multirow{-2}{*}{30 objs} & $131.5\pm5.1$  & $144.5\pm7.0$  & $151.2\pm6.5$ \\ \hline

&                           & \textbf{97\%}  & \textbf{95\%}  & \textbf{98\%} \\
& \multirow{-2}{*}{20 objs} & $80.1\pm2.3$   & $83.2\pm3.7$   & $83.7\pm3.6$  \\ \cline{3-5}
&                           & \textbf{89\%}  & \textbf{89\%}  & \textbf{82\%} \\
\multirow{-4}{*}{4 classes}
& \multirow{-2}{*}{30 objs} & $162.6\pm4.6$  & $182.7\pm6.6$  & $195.9\pm8.7$ \\ \hline

\multicolumn{2}{c}{Average Success}  & \textbf{95.2\%} & \textbf{94.7\%} & \textbf{93.7\%} \\
\multicolumn{2}{c}{Average \# Steps}    & $92.4$          & $99.4$          & $105.9$         \\
\bottomrule
\end{tabular}
\end{center}
\caption{Quantitative results of using different levels of uncertainty on sorting $20$ and $30$ cubes belonging to $2$, $3$ or $4$ classes
}
\label{table:noise}
\end{table}

\begin{figure}[htp]
\centering
\includegraphics[width = 0.95 \columnwidth]{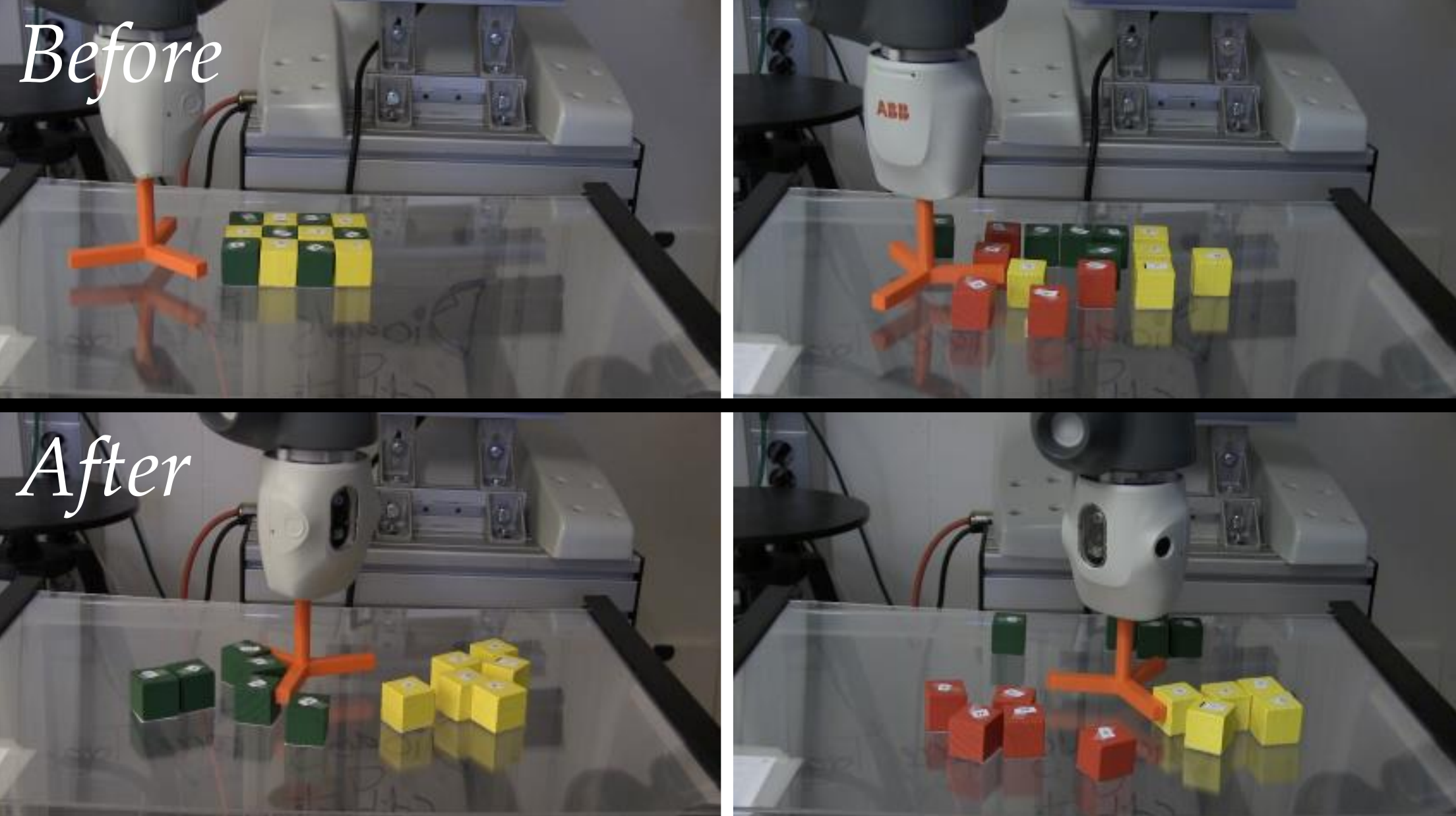}

\vspace{+0.1cm}
\footnotesize
\setlength{\tabcolsep}{4.2mm}{
\begin{tabular}{ccc}
\toprule
Test Case   & $2\times6$ cubes               & $3\times5$ cubes \\
\midrule
Simulation  & \textbf{96\%}, $31.5\pm1.2$    & \textbf{99\%}, $23.9\pm0.6$ \\
Real World  & \textbf{16/20}, $37.0\pm3.5$   & \textbf{15/20}, $27.7\pm2.1$ \\
\bottomrule
\end{tabular}
}
\caption{Quantitative real robot experiments with an ABB Yumi on two test cases.
We execute $20$ trials to sort $2 \times 6$ and $3 \times 5$ cubes respectively. For the comparison in simulation, 100 trials were run.}
\label{fig:yumi_experiments}
\end{figure}

Lastly, we run the planner on a real ABB Yumi for two test cases: sorting $2 \times 6$ cubes
and sorting $3 \times 5$ cubes. Due to hardware limitations, the workspace
in these experiments is with $39cm \times 25cm$, significantly smaller than in the previous experiments.
For each case, we run $20$ trials from the same initial configuration. The results are reported in \figref{fig:yumi_experiments}.
While in simulations we observe very few failures for these test cases, we observe more on the real robot.
All failures on the real robot occurred due to pushing a cube out of bounds.
This clearly highlights a vulnerability against uncertainty in states close to the boundary.
As long as cubes are well within the workspace, however, the closed-loop approach compensates for modeling errors.
This is reflected in the increase in the average number of actions needed to solve both cases.

\subsection{Runtime}
\label{subsec:runtime}
We parallelized the MCTS in implementation and ran it with $8$ threads on an Intel i7-7820X.
As GPU for the learned policy, we used $2\times$NVIDIA GTX 2080TI.
The learned policy network contains $11.69$M parameters, with an inference time of $2.58$ms.
For all results with random policy presented in \tableref{table:diff_cls_objs}, the average planning time per action is $2.16$s.
For the sorting tests with immovable obstacles under the 500-1500-3 setting in \figref{fig:plots}, the average planning time per action is $3.09$s with random policy and $4.37$s with learned policy.

%% file: conclusion.tex

\section{Conclusion}
\label{sec:conclusion}
We addressed a planar non-prehensile sorting task, where a robot needs to separate many objects according to user-defined class membership.
In this problem, the robot needs to disentangle, circumnavigate and simultaneously push multiple objects.
We adopted Monte Carlo tree search to solve this task and observed its effectiveness in various sorting scenarios, despite only being equipped with a random rollout policy.
In addition, we observed that we can improve the algorithm's performance by equipping it with a learned rollout policy trained from planning experiences.

These results are encouraging to further develop the use of Monte Carlo tree search for non-prehensile rearrangement.
In future work, we intend to extend the approach to minimize the number of actions needed to achieve a sorted state. Further, the heuristic reward signal could be replaced by a value network trained from sorting experience to better estimate the states.
Lastly, we believe the efficiency can yet be further improved from reusing previously grown search trees to save computation time.

%% file: root.bbl
\begin{thebibliography}{10}

\bibitem{Wilfongpspace}
G.~Wilfong, ``{Motion planning in the presence of movable obstacles},'' {\em
  Annals of Mathematics and Artificial Intelligence}, March 1991.

\bibitem{Stilman2005}
M.~Stilman and J.~J. Kuffner, ``{Navigation among movable obstacles: real-time
  reasoning in complex environments},'' {\em International Journal of Humanoid
  Robotics}, vol.~02, pp.~479--503, December 2005.

\bibitem{Stilman2008}
M.~Stilman and J.~Kuffner, ``{Planning among movable obstacles with artificial
  constraints},'' in {\em Int. Journal of Robotics Research}, Nov 2008.

\bibitem{VandenBerg2009}
J.~Van Den~Berg, M.~Stilman, J.~Kuffner, M.~Lin, and D.~Manocha, ``Path
  planning among movable obstacles: a probabilistically complete approach,'' in
  {\em Algorithmic Foundation of Robotics VIII}, Springer, 2009.

\bibitem{Nieuwenhuisen2008}
D.~Nieuwenhuisen, A.~F. van~der Stappen, and M.~H. Overmars, ``An effective
  framework for path planning amidst movable obstacles,'' in {\em Algorithmic
  Foundation of Robotics VII}, pp.~87--102, Springer, 2008.

\bibitem{Kitaev2015}
N.~Kitaev, I.~Mordatch, S.~Patil, and P.~Abbeel, ``Physics-based trajectory
  optimization for grasping in cluttered environments,'' {\em IEEE Int. Conf.
  Robotics and Automation}, May 2015.

\bibitem{Agboh2018}
W.~C. Agboh and M.~R. Dogar, ``Real-time online re-planning for grasping under
  clutter and uncertainty,'' {\em IEEE-RAS Int. Conf. Humanoid Robots}, 2018.

\bibitem{Muhayyuddin2018}
{Muhayyuddin}, M.~{Moll}, L.~{Kavraki}, and J.~{Rosell}, ``Randomized
  physics-based motion planning for grasping in cluttered and uncertain
  environments,'' {\em IEEE Robotics and Automation Letters}, April 2018.

\bibitem{Stilman2007}
M.~Stilman, J.~U. Schamburek, J.~Kuffner, and T.~Asfour, ``{Manipulation
  planning among movable obstacles},'' {\em IEEE Int. Conf. Robotics and
  Automation}, April 2007.

\bibitem{Ben-Shahar1998b}
O.~Ben-Shahar and E.~Rivlin, ``{Practical pushing planning for rearrangement
  tasks},'' {\em IEEE Transactions on Robotics and Automation}, vol.~14, no.~4,
  pp.~549--565, 1998.

\bibitem{Krontiris2015}
A.~Krontiris and K.~Bekris, ``{Dealing with Difficult Instances of Object
  Rearrangement},'' {\em Robotics: Science and Systems}, July 2015.

\bibitem{Krontiris2016}
A.~Krontiris and K.~E. Bekris, ``{Efficiently solving general rearrangement
  tasks: A fast extension primitive for an incremental sampling-based
  planner},'' {\em IEEE Int. Conf. Robotics and Automation}, May 2016.

\bibitem{Garrett2018}
C.~R. Garrett, T.~Lozano-P{\'{e}}rez, and L.~P. Kaelbling, ``{FFRob: Leveraging
  symbolic planning for efficient task and motion planning},'' {\em The
  International Journal of Robotics Research}, vol.~37, no.~1, 2018.

\bibitem{Han2017}
S.~Han, N.~Stiffler, A.~Krontiris, K.~Bekris, and J.~Yu, ``High-quality
  tabletop rearrangement with overhand grasps: Hardness results and fast
  methods,'' {\em Robotics: Science and Systems}, 2017.

\bibitem{Huang2019}
E.~Huang, Z.~Jia, and M.~T. Mason, ``Large-scale multi-object rearrangement,''
  in {\em IEEE Int. Conf. Robotics and Automation}, 2019.

\bibitem{MCTSSurvey}
C.~B. {Browne}, E.~{Powley}, D.~{Whitehouse}, S.~M. {Lucas}, P.~I. {Cowling},
  P.~{Rohlfshagen}, S.~{Tavener}, D.~{Perez}, S.~{Samothrakis}, and
  S.~{Colton}, ``A survey of monte carlo tree search methods,'' {\em IEEE
  Transactions on Computational Intelligence and AI in Games}, 2012.

\bibitem{silver2016mastering}
D.~Silver, A.~Huang, C.~J. Maddison, A.~Guez, L.~Sifre, G.~Van Den~Driessche,
  {\em et~al.}, ``Mastering the game of go with deep neural networks and tree
  search,'' {\em nature}, vol.~529, no.~7587, p.~484, 2016.

\bibitem{Lynch1996}
K.~M. Lynch and M.~T. Mason, ``{Stable pushing: Mechanics, controllability, and
  planning},'' {\em Int. Journal of Robotics Research}, 1996.

\bibitem{King2015}
J.~E. King, J.~A. Haustein, S.~S. Srinivasa, and T.~Asfour, ``{Nonprehensile
  whole arm rearrangement planning on physics manifolds},'' {\em IEEE Int.
  Conf. Robotics and Automation}, June 2015.

\bibitem{King2016}
J.~E. King, M.~Cognetti, and S.~S. Srinivasa, ``{Rearrangement planning using
  object-centric and robot-centric action spaces},'' {\em IEEE Int. Conf.
  Robotics and Automation}, May 2016.

\bibitem{King2017}
J.~E. King, V.~Ranganeni, and S.~S. Srinivasa, ``{Unobservable Monte Carlo
  planning for nonprehensile rearrangement tasks},'' {\em IEEE Int. Conf.
  Robotics and Automation}, May 2017.

\bibitem{Haustein2015}
J.~A. Haustein, J.~King, S.~S. Srinivasa, and T.~Asfour, ``{Kinodynamic
  randomized rearrangement planning via dynamic transitions between statically
  stable states},'' {\em IEEE Int. Conf. Robotics and Automation}.

\bibitem{Haustein2018}
J.~A. Haustein, I.~Arnekvist, J.~Stork, K.~Hang, and D.~Kragic, ``Learning
  manipulation states and actions for efficient non-prehensile rearrangement
  planning,'' {\em arXiv preprint arXiv:1901.03557}, 2019.

\bibitem{Bejjani2018}
W.~Bejjani, R.~Papallas, M.~Leonetti, and M.~Dogar, ``Planning with a receding
  horizon for manipulation in clutter using a learned value function,'' {\em
  IEEE-RAS Int. Conf. Humanoid Robots}, 2018.

\bibitem{Pinto2018}
L.~Pinto, A.~Mandalika, B.~Hou, and S.~S. Srinivasa, ``Sample-efficient
  learning of nonprehensile manipulation policies via physics-based informed
  state distributions,'' {\em CoRR}, vol.~abs/1810.10654, 2018.

\bibitem{Elliott2016}
S.~{Elliott}, M.~{Valente}, and M.~{Cakmak}, ``Making objects graspable in
  confined environments through push and pull manipulation with a tool,'' in
  {\em IEEE Int. Conf. Robotics and Automation}, May 2016.

\bibitem{Laskey2016}
M.~{Laskey}, J.~{Lee}, C.~{Chuck}, D.~{Gealy}, W.~{Hsieh}, F.~T. {Pokorny},
  A.~D. {Dragan}, and K.~{Goldberg}, ``Robot grasping in clutter: Using a
  hierarchy of supervisors for learning from demonstrations,'' in {\em IEEE
  Int. Conf. Automation Science and Engineering}, Aug 2016.

\bibitem{Chang2012}
L.~{Chang}, J.~R. {Smith}, and D.~{Fox}, ``Interactive singulation of objects
  from a pile,'' in {\em IEEE Int. Conf. Robotics and Automation}, May 2012.

\bibitem{Hermans2012}
T.~{Hermans}, J.~M. {Rehg}, and A.~{Bobick}, ``Guided pushing for object
  singulation,'' in {\em IEEE Int. Conf. Robotics and Automation}, Oct 2012.

\bibitem{Eitel2017}
A.~Eitel, N.~Hauff, and W.~Burgard, ``Learning to singulate objects using a
  push proposal network,'' in {\em Robotics Research}, 2020.

\bibitem{Danielczuk2018}
M.~Danielczuk, J.~Mahler, C.~Correa, and K.~Goldberg, ``Linear push policies to
  increase grasp access for robot bin picking,'' in {\em IEEE Int. Conf.
  Automation Science and Engineering}, IEEE, 2018.

\bibitem{Zagoruyko2019}
S.~Zagoruyko, Y.~Labb{\'e}, I.~Kalevatykh, I.~Laptev, J.~Carpentier, M.~Aubry,
  and J.~Sivic, ``Monte-carlo tree search for efficient visually guided
  rearrangement planning,'' {\em ArXiv}, vol.~abs/1904.10348, 2019.

\bibitem{king2017unobservable}
J.~E. King, V.~Ranganeni, and S.~S. Srinivasa, ``Unobservable monte carlo
  planning for nonprehensile rearrangement tasks,'' in {\em IEEE Int. Conf.
  Robotics and Automation}, IEEE, 2017.

\bibitem{auer2002finite}
P.~Auer, N.~Cesa-Bianchi, and P.~Fischer, ``Finite-time analysis of the
  multiarmed bandit problem,'' {\em Machine learning}, pp.~235--256, 2002.

\bibitem{He2015}
K.~He, X.~Zhang, S.~Ren, and J.~Sun, ``Deep residual learning for image
  recognition,'' {\em arXiv preprint arXiv:1512.03385}, 2015.

\end{thebibliography}
